\documentclass[conference]{IEEEtran}
\usepackage{amsmath,amssymb}
\usepackage{graphicx}
\usepackage{booktabs}
\usepackage{color}
\usepackage[hidelinks]{hyperref}
\graphicspath{{../figures/}}

\newcommand{\ostar}[1]{o^*_{#1}}
\newcommand{\ofit}{o_{\text{fit}}}
\newcommand{\obayes}{o_{\text{bayes}}}
\newcommand{\postnorm}{\textit{post\_norm}}
\newcommand{\fitcost}{\textit{fit\_cost}}

\begin{document}

\title{Robust Is Salient: An Informed Adversary Moves the Optimal Signal onto the Salience Pole}
\author{
\IEEEauthorblockN{Cris Huynh}
\IEEEauthorblockA{Independent researcher\\crishuynh2004@gmail.com}
}
\maketitle

\begin{abstract}
When an informed adversary shares the audience of a constrained signalling
channel, the signal that best protects the truth is the signal that best
describes it. On every one of 108 confirmatory items 
the adversary-robust optimum is the salience pole of Paper 1, and across a
200{,}000-item candidate pool 
all 2{,}748 items where the two differ 
lie where Paper 1's salience-to-Bayes coordinate is undefined. 
Wherever that coordinate is defined, robustness is bought by walking back the
whole interval from Bayesian discrimination to salience. We obtain this by adding to Paper 1's
forced-choice task, abstracted from \emph{Deception: Murder in Hong Kong}, an
adversary who knows the target, sees the signal, and argues for the strongest
wrong answer with persuasion budget $\beta$. The optimal signal moves from the
posterior-maximizing option to the margin-maximizing one as $\beta$ grows, and
at $\beta = 0$ the game reproduces Paper 1's oracle exactly, at the listener
temperature $\tau = 1$ inherited from Paper 1's frozen listener. The effect is
real: 18.2 percent of the pool 
has an optimum some finite budget moves, and each item's critical budget is
exact. The coincidence fixes what an evaluation can learn. Two adversary
framings change the chosen option of seven language models 
on 30 to 77 of 108 
items against an exact no-effect rate, and no measurement on these items can
say whether that movement is toward the adversary-aware optimum or toward
salience, because the two are one option. This is a structural limit, not a
null result. The check that exposes it is cheap: before evaluating
adversary-awareness, test whether the robust target coincides with a heuristic
target on the evaluation items.
\end{abstract}

\section{Introduction}
Paper 1 \cite{huynh2026consistency} measured where language models land in a
forced-choice signalling task whose optimum is computable in closed form. One
party knows a hidden target pair of concepts and may send exactly one option
from a small, fixed menu. Two references bound each item: the salience pole,
the option that fits the target best, and the Bayes pole, the option that
maximizes a literal listener's posterior on the target. Every model tested was
item-sensitive, and every model sat outside the salience-to-Bayes interval on
the far side of salience, some of them indistinguishable from uniform random
selection. None approached the option a listener-modelling speaker would send.

Real audiences are not always cooperative. In the board game the task is
abstracted from, the speaker shares the table with a player who knows the
answer and wants the others to guess wrong. This paper adds that player. The
adversary sees the signal, picks the wrong answer the signal leaves most
plausible, and argues for it with persuasion budget $\beta$. Against that
adversary the best signal is no longer the one that maximizes the posterior on
the truth. It is the one that maximizes the truth's margin over its strongest
rival, and the objective moves continuously from the first to the second as
$\beta$ grows. At $\beta = 0$ the adversary game reproduces Paper 1's oracle
exactly. That nesting holds at listener temperature $\tau = 1$, which Paper 2
does not choose: it is the temperature at which Paper 1's frozen listener has
Paper 1's oracle as its optimum at all.

The effect exists and is computable. Every item has an exact critical budget
$\beta_c$ at which its optimum first moves, in closed form. On 18.2 percent of
Paper 1's 200{,}000 candidate items 
some finite budget moves the optimum, and on the items where it does, the
speaker gives up a measurable share of the listener's posterior on the truth to
buy margin.

The central finding is where the robust signal goes. On all 108 items of the
confirmatory set, 
the adversary-robust optimum is Paper 1's salience pole, the option that
simply fits the target best. Across the pool the two differ on 2{,}748 of
36{,}464 divergent items, 
and every one of those lies where Paper 1's salience-to-Bayes coordinate is
undefined. Wherever that coordinate exists, robustness against an informed
adversary is bought by giving up the whole interval from Bayesian
discrimination back to salience. The two targets separate only where the
coordinate cannot see them.

That has a direct consequence for evaluation. We ran seven language models 
under two framings that describe the adversary. The framings changed the
chosen option on every model, against a no-effect rate that is exactly 1 by
construction of the scorer. What that movement was toward cannot be measured
on this item set, by any statistic, because on every confirmatory item the
adversary-aware target and the salience target are the same option. A model
that moved to it and a model that drifted to salience make the identical
choice. We report this as a finding about the evaluation, not as a null: the
experiment did not look for a direction effect and fail to find one, it could
not have found one.

The lesson we draw is narrow. Before evaluating whether models are
adversary-aware, check whether the adversary-robust target coincides with a
heuristic target on the evaluation items. On this task the check was
necessary, and it is cheap: it needs no model output, only the oracle
computations the evaluation already requires.

\paragraph{Contributions}
\begin{enumerate}
\item A formalization of an informed adversary sharing the audience, with the
adversary's best response and the speaker's objective in closed form, and an
exact nesting of Paper 1's oracle at $\beta = 0$ under the inherited condition
$\tau = 1$ (Section~\ref{sec:setup}).
\item A characterization of the effect on Paper 1's frozen geometry: the
divergence set on the full pool, the exact critical budget $\beta_c$, and the
price of robustness on Paper 1's own normalization (Section~\ref{sec:arma}).
\item The coincidence: the adversary-robust optimum is Paper 1's salience pole
on every confirmatory item, and the two separate only where Paper 1's frontier
coordinate is undefined (Section~\ref{sec:coincidence}).
\item An evaluation of seven models under two adversary framings, which
establishes that the framings move choices against an exact zero, and a
structural account of why the direction of that movement is unanswerable on
this item set (Section~\ref{sec:armb}).
\end{enumerate}

\section{Related Work}\label{sec:related}
\paragraph{Adversary-aware speakers, and what can be measured}
The closest prior work is the Stackelberg Speaker \cite{zheng2025stackelberg},
which formalizes turn-based dialogue in social deduction games as a Stackelberg
competition and trains speakers by reinforcement learning to optimize the
persuasive impact of their utterances across three games. Its utterances are
free-form, so its setting admits no ground-truth optimal message, and its
evaluation is necessarily relative: win rates and comparisons against baselines
and ablations. Our task is built so that the missing quantity exists. The signal
space is small and enumerable, the listener is a parameter-free literal Bayesian
model inherited frozen from Paper 1, and the adversary's best response is
solved in closed form. For every item and every adversary strength the optimal
signal is exactly computable, so a choice's distance from optimal is a number
rather than a comparison. The contribution is measurability, not the
observation that adversaries change what a speaker should say.

\paragraph{Codenames: a fixed hazard against a responsive adversary}
Codenames is the nearest structural analogue: a sender picks one signal from a
constrained space to point a receiver at a target set. It has been proposed as
a language-model benchmark \cite{stephenson2024codenames}, modelled with
Bayesian inference over partner types drawn from a cognitive hierarchy
\cite{bills2024cooperation}, and formulated as a Markov decision process on
which standard reinforcement-learning algorithms fail to converge
\cite{siu2022codenames}. In every treatment the assassin is a board tile fixed
before the clue is given, a hazard whose location is known and static. Our
adversary is a player: it observes the signal and only then chooses which
rival hypothesis to promote. A fixed hazard can be routed around. A responsive
one cannot, because whatever ambiguity a signal leaves is what the adversary
will occupy, and that change turns a constrained optimization into a game.

\paragraph{Information design with competing senders}
Bayesian persuasion \cite{kamenica2011persuasion} and its competitive extensions
\cite{gentzkow2017competition,gentzkow2017rich} already contain this paper's
normative question in abstract form. In the zero-sum case most sender
preferences yield full revelation \cite{ravindran2020zerosum}. Our speaker
cannot fully reveal: each tile offers three to six options, 
so the binding force is the channel, not the preference structure. Computing an
equilibrium in multi-sender persuasion is PPAD-hard, and a single sender's best
response is NP-hard \cite{hossain2024multisender}. That hardness is why a small,
exactly solvable, semantic instance is worth having: it is a place to check
what the approximations the general problem forces are approximating.

\paragraph{Evaluating language models in social deduction}
Two recent evaluations diagnose the problem our setup addresses. MINDGAMES
\cite{wang2026mindgames} reports that failure-heavy environments can reward
robustness to opponent errors as much as strategic ability, with an
error-survival confound in Secret Mafia, the environment closest to our source
game. QUACK \cite{yuan2026quack} argues that outcome-only scoring cannot tell
whether an agent's language is grounded in what it perceived, and audits
claims against engine logs. Neither supplies a reference optimum, because their
settings do not admit one. Section~\ref{sec:armb} shows that having one is
necessary and not sufficient: an exact optimum still cannot score direction
when it coincides with a heuristic target.

\paragraph{The human-behaviour gap}
Speech-production research frames the listener as someone to help
\cite{hazan2011acoustic,buz2016dynamically,gessa2026louder}. Experimental
economics has studied competitive context effects on senders, but measures
honesty rate under misaligned incentives
\cite{rode2010truth,sutter2009deception,gneezy2005deception,cai2006overcommunication}.
Neither asks which signal a truthful sender picks when a hostile co-speaker will
exploit residual ambiguity. Pinker, Nowak and Lee \cite{pinker2008indirect} are
the closest bridge, but concern semantic indirectness, not signal selection
against a computable optimum. This paper has no human arm and makes no claim
about what human speakers do.

\section{Setup and the Adversary Game}\label{sec:setup}
\paragraph{Paper 1's task}
An item shows $M = C = 10$ candidate Means and Clues, so the hypothesis space
$H$ is the $|H| = 100$ pairings, 
exactly one of which is the target $h^*$. The speaker sees one tile, a
categorical dimension with option set $O$ of 3, 3, 4 or 6 options on
\texttt{manmade}, \texttt{moves}, \texttt{hold} and \texttt{size}. 
The fit $f(o,h)$ of option $o$ to hypothesis $h$ is Paper 1's min-aggregated
exponential decay with no free parameter, and the prior is uniform. The literal
listener's posterior is
\begin{equation}
L(h \mid o) = \frac{P(o\mid h)}{\sum_{h'} P(o \mid h')},\qquad
P(o\mid h) = \frac{f(o,h)}{\sum_{o'} f(o',h)},
\end{equation}
which is the posterior Paper 1's frozen oracle was computed with.%
\footnote{Paper 1's printed Eq.~(2) writes $f(o,h^*)/\sum_h f(o,h)$ and omits
the normalization over options. The frozen oracle and item set were built with
it, and every quantity here uses the form the artifact was built with.} 
Paper 1's two poles are
$\ofit = \arg\max_o f(o,h^*)$ and $\obayes = \arg\max_o L(h^*\mid o)$, ties
broken by Paper 1's rule (maximum fit among options tied on the posterior).
Paper 1 places a chosen option $a$ on
$\postnorm(a)$, the posterior on $h^*$ min-max normalized over the item's own
options, and on the salience-to-Bayes coordinate
$\textit{sb}(a) = (\postnorm(a) - \postnorm(\ofit)) / (1 - \postnorm(\ofit))$,
which is 0 at the salience pole and 1 at the Bayes pole and is undefined where
the span $1 - \postnorm(\ofit)$ is at or below Paper 1's guard of 0.02. 

\paragraph{The adversary}
The adversary knows $h^*$, observes $o$, and promotes one decoy $d \ne h^*$ with
persuasion budget $\beta \ge 0$. Honest investigators score
$u(h) = \log L(h\mid o) + \beta\,\mathbb{1}[h = d]$ and choose by quantal
response at temperature $\tau$. The speaker's value is the adversary-worst-case
probability of a correct accusation,
$V_\beta(o) = \min_{d \ne h^*} P(\text{choose } h^* \mid o, d)$. The listener
is Paper 1's, frozen; $\beta$ is the one new parameter.

\paragraph{R1: the adversary's best response}
Substituting $u$ shows $P(h^*\mid o, d)$ is strictly decreasing in $L(d\mid o)$
for every $\beta \ge 0$ and $\tau > 0$, so
\begin{equation}
d^*(o) = \arg\max_{h \ne h^*} L(h \mid o),
\end{equation}
the strongest surviving rival under the signal actually sent, independent of
$\beta$ and $\tau$.

\paragraph{R2: the $\beta$ continuum}
Writing $M(o) = L(d^*(o)\mid o)$,
\begin{equation}
V_\beta(o) = \frac{L(h^*\mid o)^{1/\tau}}
{\sum_h L(h\mid o)^{1/\tau} + M(o)^{1/\tau}\,(e^{\beta/\tau} - 1)}.
\end{equation}
As $\beta \to \infty$ the budget term dominates and, for every $\tau > 0$,
$\arg\max_o V_\beta(o) \to \arg\max_o \textit{margin}(o) = \ostar{\infty}$, with
$\textit{margin}(o) = \log L(h^*\mid o) - \log M(o)$.

\paragraph{The nesting, and why $\tau = 1$}
At $\beta = 0$ the denominator is $\sum_h L(h\mid o)^{1/\tau}$. Because
$L(\cdot\mid o)$ is a distribution over $H$, that sum is identically 1 at
$\tau = 1$, so $V_0(o) = L(h^*\mid o)$ and
$\ostar{0} = \arg\max_o V_0(o) = \obayes$, Paper 1's oracle, exactly. The
identity holds only at $\tau = 1$: at any other temperature the denominator is
a power sum that depends on the shape of $L(\cdot\mid o)$, and a
counterexample exists. We do not choose $\tau = 1$. Paper 1's oracle is the
optimal signal for a $\tau = 1$ quantal-response listener and for no other, so
$\tau = 1$ is inherited from Paper 1's frozen listener; any other value would
change which option is the no-adversary optimum, which is a change to a frozen
Paper 1 artifact. The condition was found by re-deriving R2, whose first draft
assumed it without stating it. Every claim in this paper that ties the
adversary game to Paper 1 holds at $\tau = 1$, and we claim no robustness
to $\tau$.

\paragraph{The critical budget $\beta_c$}
For fixed $o$, $V_\beta(o) = C_o / (A_o + B_o x)$ with $x = e^{\beta/\tau}$ and
constants $A_o, B_o, C_o > 0$. Two options' values are equal where a linear
equation in $x$ holds, so each rival overtakes $\ostar{0}$ at most once and
never gives the lead back. An item's critical budget is therefore
$\beta_c(i) = \min_{o' \ne \ostar{0}} \beta^*_{o'}$, infinite if no rival ever
overtakes, and the divergence sets $D(\beta) = \{i : \beta_c(i) \le \beta\}$
are nested in $\beta$, with $D(0)$ empty by Paper 1's tie-break. The single
crossing holds at any $\tau$; what needs $\tau = 1$ is that $\beta_c$ is
measured from Paper 1's oracle. Ties in $\arg\max_o V_\beta(o)$ are broken as
Paper 1 breaks ties in $\obayes$. The value of record is the bisected
$\beta_c$; the closed-form root is a cross-check, and it agrees with bisection
on every item of Paper 1's frozen 1{,}000 (Appendix~\ref{app:derivations}). 

\paragraph{Two measures}
The \emph{divergence set} is $D(\infty)$, the items on which some finite budget
moves the optimum; it is empty exactly when $\ostar{0} = \ostar{\infty}$. The
\emph{price of robustness} is $L(h^*\mid\ostar{0}) - L(h^*\mid\ostar{\infty})$,
the posterior mass on the truth given up to buy margin, reported primarily on
Paper 1's $\postnorm$ scale, where it is
$\postnorm(\ostar{0}) - \postnorm(\ostar{\infty})$.

\section{Arm A: The Adversary Effect}\label{sec:arma}
\paragraph{The effect exists in the signal space}
On the full pool of 200{,}000 candidates, 
18.2 percent of items have a finite $\beta_c$, 
against a preregistered kill threshold of 10 percent, 
so the kill gate passes. 
The rate on the reporting grid's endpoint, $|D(8)|/N$, is 18.0 percent. 
Of the 36{,}464 divergent pool items, 
136 have a $\beta_c$ set by Paper 1's tie-break band rather than by a strict crossing. 
All of them lie between $\beta = 17.35$ and $23.45$, above the grid endpoint, 
so they enter $|D(\infty)|$ and no grid-rate figure; none of them separates the
robust optimum from salience, 
and the frozen 1{,}000 contains none (Appendix~\ref{app:tolerance}). 
The 18.2 percent is the population figure. 
Paper 1's frozen 1{,}000 
diverge at 46.0 percent, 
a property of that set's 50 percent conflict quota rather than of the signal
space, 
and we never quote it as a population rate.

\begin{figure}[t]
\centering
\includegraphics[width=\linewidth]{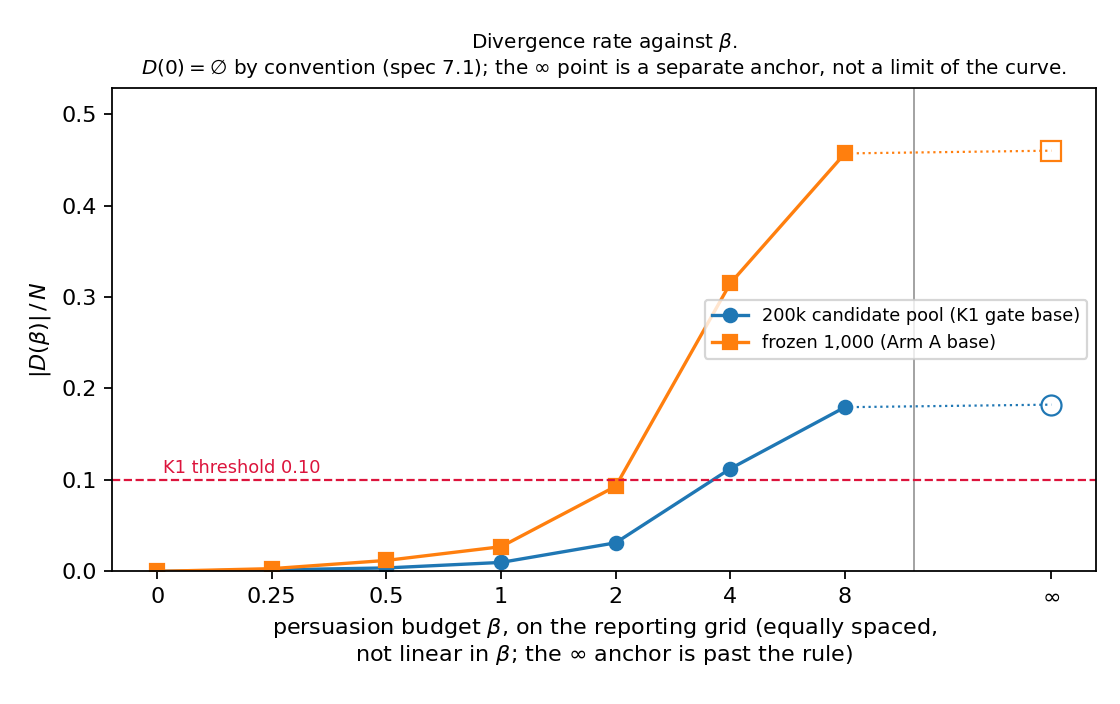}
\caption{Divergence rate $|D(\beta)|/N$ on the reporting grid, for the 200{,}000-item 
pool, which is the population base, and for Paper 1's frozen 1{,}000, whose 
rate reflects its conflict quota. $D(0)$ is empty by construction and the
$\infty$ point is a separate anchor, not a limit of the curve. The dashed line
is the preregistered kill threshold of 0.10.} 
\label{fig:divergence}
\end{figure}

\paragraph{Divergence grows with the budget}
The pool's divergence rate rises strictly across the reporting grid, from 0.16
percent at $\beta = 0.25$ to 3.1 percent at $\beta = 2$ and 18.0 percent at
$\beta = 8$ (Fig.~\ref{fig:divergence}). 
No item has $\beta_c = 0$. 
By tile the existence rate runs from 3.9 percent on \texttt{manmade} to 33.4
percent on \texttt{size}, 
and it is not monotone in option count: \texttt{moves}, with three options,
diverges at 17.9 percent, above \texttt{hold}'s 17.7 percent with four. 

\begin{figure}[t]
\centering
\includegraphics[width=\linewidth]{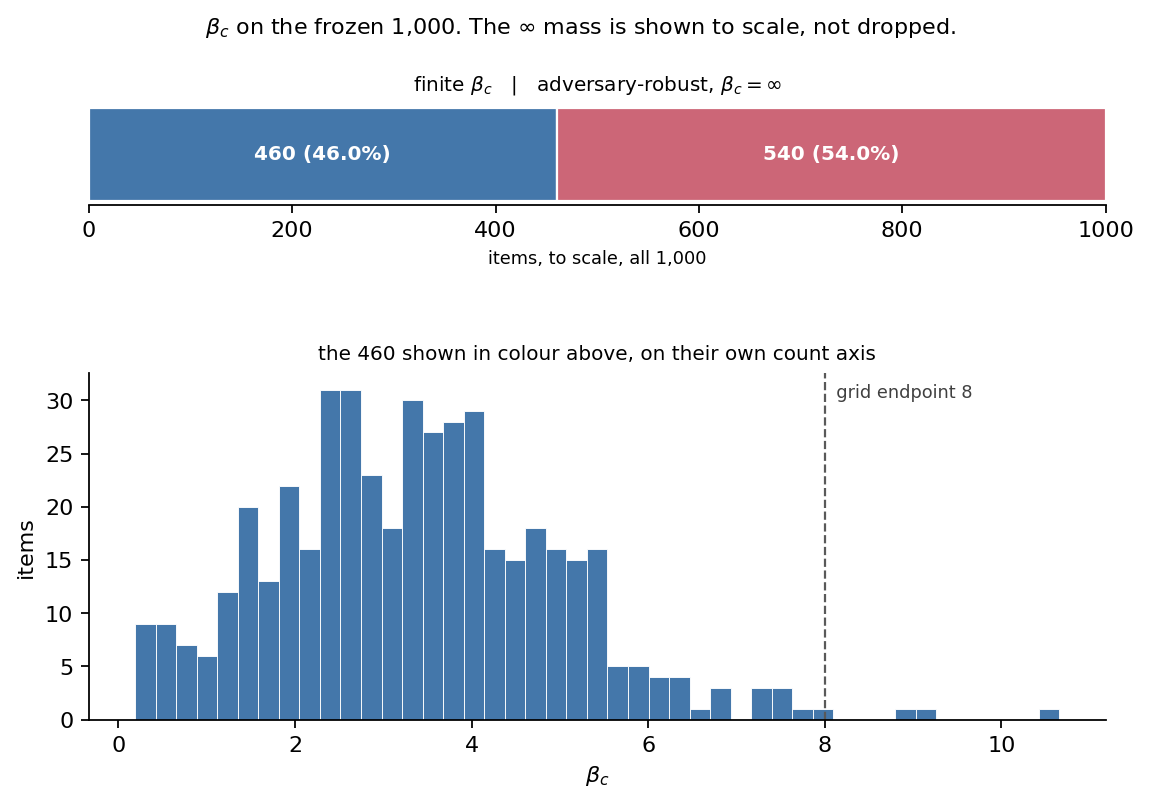}
\caption{$\beta_c$ on Paper 1's frozen 1{,}000. 
Top: the 460 items with finite
$\beta_c$ against the 540 adversary-robust items, to scale. Bottom: the finite
values.} 
\label{fig:betac}
\end{figure}

\paragraph{The critical budget is an exact per-item scalar}
On the 460 frozen items with finite $\beta_c$, 
the median critical budget is 3.33 and the mean 3.36 (SD 1.61), 
from 0.19 to 10.66 (Fig.~\ref{fig:betac}). 
Three lie above the grid endpoint of 8. 
The remaining 540 frozen items are adversary-robust: no budget moves their
optimum. 
What makes an item cheap to attack is exploratory, with no preregistered
prediction and no test. 
With that stated: $\beta_c$ falls as Paper 1's \fitcost{} rises, Spearman
$-0.57$ on the finite subset, 
and tiles differ 8.6-fold in whether items diverge but only 1.22-fold in the
median budget needed once they do. 
Appendix~\ref{app:arma} gives the full exploratory analysis.

\begin{figure}[t]
\centering
\includegraphics[width=\linewidth]{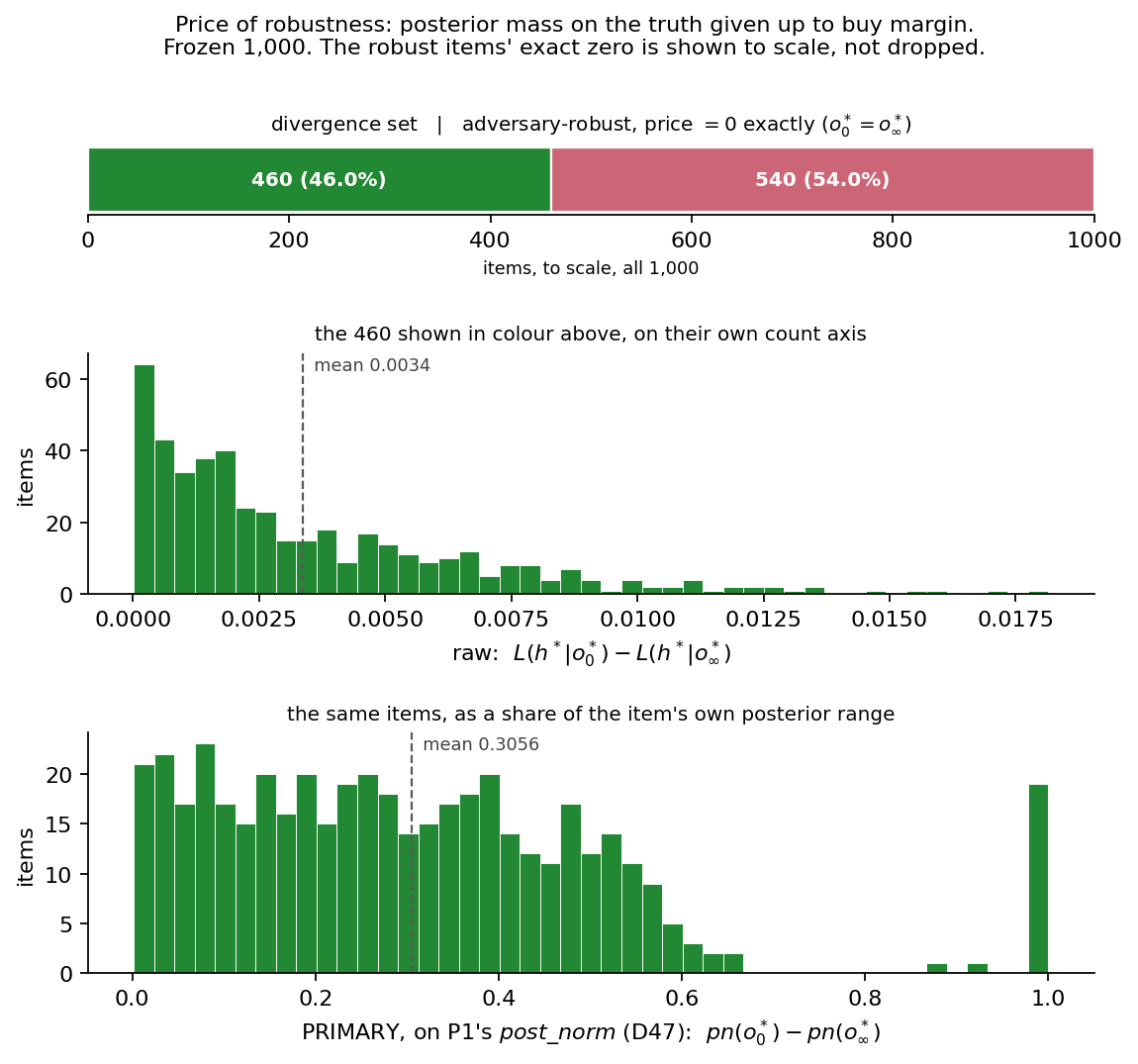}
\caption{Price of robustness on Paper 1's frozen 1{,}000. The 540 robust items 
pay exactly zero. 
Middle: raw posterior mass on the truth given up. Bottom: the
same items on Paper 1's $\postnorm$ scale, the primary convention; the spike at
1 is the nineteen items whose robust signal is their lowest-posterior option.} 
\label{fig:price}
\end{figure}

\paragraph{Robustness has a price on Paper 1's own scale}
Across the 36{,}464 divergent pool items, the adversary-robust signal gives up 
on average 0.276 of the item's posterior range on the truth. 
That is the population figure. On the 460 frozen divergent items the same mean
is 0.306, the median 0.273, and the value from the means 0.298, 
defined on all 460. 
We report the median and the value from the means beside the mean because the
quantity is a mean of per-item ratios, the pathology Paper 1 measured directly.
On the raw probability scale the price averages 0.0034 (SD 0.0033). 
Nineteen items sit at exactly 1, where the robust signal is the item's
lowest-posterior option; all nineteen are $\ofit$, and seventeen are on
\texttt{manmade}. 
We keep them: they are a corner of the signal space, not a degeneracy. The
price is never negative, which would be an implementation bug since $\ostar{0}$
is the posterior argmax. 
Fig.~\ref{fig:price} shows both scales, and Appendix~\ref{app:arma} gives the
two alternative denominators.

\paragraph{The preregistered \fitcost{} gate fired}
The design planned to redraw Arm B's items stratified on $\beta_c$, gated on
$\beta_c$ not tracking Paper 1's \fitcost{} too closely. On the pool the
Spearman correlation is $-0.9036$, beyond the threshold of 0.8 in magnitude,
so the gate fired. 
The reason is a coincidence of three optima: where \fitcost{} is zero,
$\ofit = \obayes$ and 98.4 percent of frozen items are adversary-robust,
against 9.6 percent where \fitcost{} is positive. 
The redraw did not happen and Arm B runs on Paper 1's frozen items, which is
where its $n = 108$ comes from (Section~\ref{sec:discussion}).

\section{The Coincidence}\label{sec:coincidence}
\paragraph{The adversary-robust signal is the salient signal}
On every one of the 108 confirmatory items, the option that maximizes the
truth's margin over its strongest rival is the option that fits the truth
best: $\ostar{\infty} = \ofit$ on 108 of 108. 
The two are different functions. One ranks options by how well they separate
$h^*$ from the rival the adversary will promote; the other ranks them by how
well they describe $h^*$, ignoring every rival. Nothing in the definitions
makes them agree, and on other tiles they do not always agree. On
\texttt{size} they agree on every item where the adversary moves the optimum at
all. The measurement uses only Paper 1's frozen, model-free geometry. It reads
no model output.

\paragraph{Three bases, three figures}
The coincidence is stated separately on each base, because the bases differ and
a figure on one understates or overstates another.
\begin{itemize}
\item \emph{Confirmatory set, 108 items.} $\ostar{\infty} = \ofit$ on 108 of 108. 
None of the separating items in the frozen set is on \texttt{size}. 
On the 101 confirmatory items above the adversary-extent floor, the salience
reference sits at the adversary oracle's value on all 101. 
\item \emph{Frozen divergence set, 460 items, all four tiles.} $\ostar{\infty}
= \ofit$ on 452 of 460. 
The 8 exceptions are 3 on \texttt{manmade} and 5 on \texttt{moves}, 
and all 8 fall inside Paper 1's span guard, which removes 21 of the 460. 
On all 439 items where Paper 1's coordinate is defined, the robust signal sits
at $\textit{sb} = 0$, exactly on the salience pole. 
\item \emph{Pool, 36{,}464 divergent items.} The two differ on 2{,}748, 
and all 2{,}748 fall inside the span guard, 
which removes 4{,}010 pool items in all. 
On the 32{,}454 where the coordinate is defined, the robust signal sits on the
salience pole on every one. 
The count of 2{,}748 is the same under both methods of computing $\beta_c$, 
and none of the 2{,}748 is among the tolerance-determined items. 
\end{itemize}

\paragraph{Robustness is bought by walking back to salience}
Measured on Paper 1's salience-to-Bayes coordinate, the price of robustness is
exactly 1 wherever the coordinate is defined, to within $6\times10^{-15}$ on
the frozen set and $7\times10^{-14}$ on the pool. 
The speaker facing an informed adversary gives up the entire interval between
Bayesian discrimination and salience. It does not move partway toward the
salience pole; it lands on it. That is also why this coordinate cannot serve as
the price's denominator: it carries no distributional information about the
quantity, which is a constant.

\paragraph{Where the two targets separate, the coordinate is blind}
The separating items exist, and they are counted. But every one of them, on
the frozen set and on the pool, lies where the gap between $\ofit$ and $\obayes$
on Paper 1's $\postnorm$ is at or below Paper 1's guard of 0.02, 
which is to say where the salience and Bayes poles are too close for Paper 1's
coordinate to be defined. The adversary-aware target differs from the salience
target only where there is no interval to place it in. In this signal space as
measured, the separation and the coordinate do not co-occur.

\paragraph{What the coincidence is, and what it is not}
It is not an identity: the exceptions exist, and on \texttt{manmade} and
\texttt{moves} they are real items. It is a property of where the normative
optimum moves inside Paper 1's interval once an adversary shares the audience.
It sits beside Paper 1's finding that models land on the far side of the
salience pole, and it is not the same claim: nothing in this section measures a
model. Its force is on what can be measured next. If the adversary-aware
optimum is the salience pole, a model that has learned to anticipate the
adversary and a model that has never modelled a listener at all send the same
signal. Section~\ref{sec:armb} shows what that does to an evaluation.

\section{Arm B: What the Evaluation Shows, and Why It Cannot Show More}\label{sec:armb}
\subsection{Design}
\paragraph{Framings}
Three prompts differ only in how the adversary is described (Appendix~\ref{app:framings}).
$F_0$ is Paper 1's Format V prompt, byte for byte. $F_1$ adds, in the slot
before the final question, that a third person who knows the pair will speak
to the listener and wants them to name a different pair. $F_2$ shares $F_1$'s
first two sentences and replaces its third with the adversary's rule: they will
argue for whichever wrong pair the description makes look most likely. Neither
block names an option, a concept, a tile, or which signal to send.

\paragraph{Models and measurement}
The seven models are Paper 1's: 
 Qwen3 at 0.6B, 1.7B, 4B and 8B (L1 to L4), the
1.7B and 8B base models (B2, B4), and OLMo-2-7B-Instruct as the cross-family
control (CTRL). Each chooses by argmax over teacher-forced log-probabilities of
the option strings, PMI-normalized, with Paper 1's harness; there is no
sampling anywhere in the scoring path. 
\paragraph{Item sets}
The confirmatory set is the \texttt{size} tile's 108 divergent items, 
each with two stored option-order permutations. The control set is the same
tile's 142 adversary-robust items, 
where $\ostar{0} = \ostar{\infty}$ and there is nothing for an adversary-aware
model to move toward.

\paragraph{The adversary coordinate}
For a chosen option $a$ on item $i$, Paper 2's primary coordinate is
\begin{equation}
A_i(a) = \frac{\textit{mn}_i(a) - \textit{mn}_i(\ostar{0})}{1 - \textit{mn}_i(\ostar{0})},
\end{equation}
where $\textit{mn}_i$ is the margin min-max normalized over the item's options.
$A = 0$ at Paper 1's oracle and $A = 1$ at the adversary-aware optimum. On the
confirmatory set $\ostar{\infty} = \ofit$, so $A = 1$ is also the salience pole.
No chosen option in 4{,}532 renderings exceeds $A = 1$, 
which is structural: $A \le 1$ identically, and a value above it would be a bug.

\paragraph{Three quantities, none gating another}
(a) \emph{Inertness}: the share of rendering pairs whose chosen option differs
between a framing and $F_0$, against zero. (b) \emph{Magnitude context}: the
framing's same-option rate minus the same-option rate of Paper 1's
condition-5 sentence, $R_m$, a content-neutral insertion in the same slot that
says nothing about an adversary. (c) \emph{Direction}: an exact two-sided sign
test, at the item level, of the per-item change in $A$ on items where it is
nonzero, against $p_0 = 0.5$ at $\alpha = 0.05/21$. 

\subsection{The framings changed the chosen option}
\paragraph{The null is exactly zero, and was measured}
Because the scorer is a deterministic argmax, an identical prompt returns an
identical choice, and the no-effect same-option rate is exactly 1. 
We measured it. Re-running $F_0$ reproduces Paper 1's choices on all 3{,}500
confirmatory-tile renderings, with 0 disagreements: 
choice-level output identity on the confirmatory tile, with environment
identity not established. 
Off that tile the rate is nonzero, from 1 to 16 disagreeing renderings of about
1{,}500 per model, beside Paper 1's own batch-flip rates. 
Quantity (a) carries a floor of 7 items, a numerical noise floor for
floating-point near-ties and not a significance threshold; the measured
disagreement count of 0 left it unrevised, since the rule only moves it
upward. 
The zero measured on $F_0$ is a lower bound on the noise under the longer $F_1$
and $F_2$ prompts, and we report it as one. 

\paragraph{Movement on all fourteen cells} 
Both framings changed the chosen option for every model. The number of items
with at least one changed rendering pair runs from 30 of 108 (L1 under $F_1$)
to 77 of 108 (B4 under $F_2$), 
every cell clears the floor, and every bootstrap interval excludes zero
(Table~\ref{tab:armb}). 
The item denominator is 108 on every cell: at most 2 rendering pairs per cell
were excluded for a tied score, and no item lost both. 
Parse failures are structurally zero, because nothing is generated and so
nothing is parsed. 
This claim reads chosen options only. It says the adversary framings changed
what the models chose. It says nothing about why.

\begin{table*}[t]
\centering
\caption{Arm B on the 108-item \texttt{size} confirmatory set, per model and framing, PMI rule. (a): items with a changed rendering pair against $F_0$, of 108; every cell clears the floor of 7. Control: the same instrument on the 142 adversary-robust items of the same tile, exploratory (P2-D27); reported beside (a) and not comparable to it as an attribution. (b): framing same-option rate minus $R_m$, item means, so a negative value means the framing changed more choices than condition 5 did; $^\dagger$ marks the three cells whose interval excludes zero. (c): effective $n$, share of items with $\Delta A > 0$, exact two-sided $p$, tie rate; $^\ast$ marks $p < 0.05/21$. Gap: $0.5$ minus the model's content-neutral sign proportion; negative on B2.}
\label{tab:armb}
\begin{tabular}{llcccccccc}
\toprule
 & & (a) & control & (b) & \multicolumn{4}{c}{(c) direction} & \\
\cmidrule(lr){6-9}
model & framing & items /108 & items /142 & $\Delta$ vs $R_m$ & $n_{\text{eff}}$ & share $>0$ & $p$ & ties & gap \\
\midrule
CTRL & $F_1$ & 43 & 57 & $+0.065$ & 36 & 0.167 & $7.0\times10^{-5}$$^\ast$ & 0.667 & $+0.305$ \\
B2   & $F_1$ & 51 & 58 & $-0.065$ & 50 & 0.440 & 0.48 & 0.537 & $-0.056$ \\
B4   & $F_1$ & 66 & 96 & $-0.023$ & 65 & 0.477 & 0.80 & 0.398 & $+0.040$ \\
L1   & $F_1$ & 30 & 45 & $-0.023$ & 24 & 0.292 & 0.064 & 0.778 & $+0.093$ \\
L2   & $F_1$ & 42 & 41 & $-0.060$ & 42 & 0.524 & 0.88 & 0.611 & $+0.016$ \\
L3   & $F_1$ & 53 & 63 & $-0.106^\dagger$ & 47 & 0.191 & $2.5\times10^{-5}$$^\ast$ & 0.565 & $+0.258$ \\
L4   & $F_1$ & 48 & 74 & $+0.000$ & 46 & 0.217 & $1.6\times10^{-4}$$^\ast$ & 0.574 & $+0.198$ \\
\midrule
CTRL & $F_2$ & 50 & 56 & $+0.046$ & 36 & 0.361 & 0.13 & 0.667 & $+0.305$ \\
B2   & $F_2$ & 64 & 71 & $-0.148^\dagger$ & 62 & 0.339 & 0.015 & 0.426 & $-0.056$ \\
B4   & $F_2$ & 77 & 104 & $-0.102$ & 74 & 0.473 & 0.73 & 0.315 & $+0.040$ \\
L1   & $F_2$ & 32 & 46 & $-0.032$ & 28 & 0.250 & 0.013 & 0.741 & $+0.093$ \\
L2   & $F_2$ & 40 & 41 & $-0.042$ & 40 & 0.500 & 1.00 & 0.630 & $+0.016$ \\
L3   & $F_2$ & 63 & 73 & $-0.162^\dagger$ & 57 & 0.246 & $1.5\times10^{-4}$$^\ast$ & 0.472 & $+0.258$ \\
L4   & $F_2$ & 62 & 82 & $-0.097$ & 61 & 0.213 & $7.7\times10^{-6}$$^\ast$ & 0.435 & $+0.198$ \\
\bottomrule
\end{tabular}
\end{table*}

\paragraph{Magnitude context resolves on three cells and not on eleven}
Paper 1's condition-5 sentence is an active comparator, not a baseline: on
these items it changes the chosen option on 0.139 to 0.370 of rendering pairs,
with every interval excluding the no-effect rate, on all seven models. 
Against it, the framings moved choices more than condition 5 did on three
cells, L3 under $F_1$ and B2 and L3 under $F_2$, by 0.106, 0.148 and 0.162 in
same-option rate. 
On the other eleven the difference does not resolve, 
and the instrument's half-width of 0.14 to 0.17, fixed before the run, 
means a gap below about 0.15 is not resolvable. 
On those eleven cells the
framing's effect cannot be told apart from that of a content-neutral insertion
in the same slot.

\paragraph{The control set moves too}
On the 142 adversary-robust items, where the optimum does not change, the 
framings changed the chosen option on 41 to 104 items per cell. 
This change rate is exploratory: it was authorized after (a) was published,
with (a)'s instrument unmodified, and it carries no floor and no verdict. 
The control change rate and quantity (a) are reported side by side in the same
units and are not comparable as an attribution: they are taken on disjoint item
sets that differ by construction in whether the adversary can move the optimum,
no preregistered rule transfers a rate from one to the other, and no
difference, ratio or share of the two is reported. 
The preregistered control measure, the total variation distance between the
option marginals under a framing and under $F_0$, runs from 0.018 to 0.278
across cells on these 142 items, 
and is reported beside its 540-item all-tile counterpart in
Appendix~\ref{app:tables}. 
On these items any change in the choice distribution is prompt sensitivity.
The control shows that generic movement exists; it cannot size how much of the
confirmatory movement that movement explains, and no quantity in the design
can. 

\subsection{Quantity (c), reported as computed}
Five of fourteen cells resolve at the corrected $\alpha$: 
CTRL, L3 and L4 under
$F_1$, and L3 and L4 under $F_2$. 
On all five the share of moving items with $\Delta A > 0$ is below $p_0 = 0.5$,
from 0.167 to 0.246, 
at effective $n$ of 36 to 61 and tie rates of 0.435 to 0.667 (Table~\ref{tab:armb}). 
Every verdict carries its signed gap between $p_0$ and the same model's
content-neutral sign proportion. A content-neutral insertion does not drift
toward the salience pole: its item-level sign proportion runs from 0.195 to
0.556 across the seven models, and resolves at the corrected $\alpha$ on CTRL
alone, below 0.5. 
B2 sits at the null, at 0.556 on an effective $n$ of 36 with $p = 0.62$ and an
exact interval of $[0.381, 0.721]$. 
On B2 the gap is therefore negative, $-0.056$, 
which is not a smaller Type~II cost but a Type~I exposure: a B2 result
significant against 0.5 could sit at or below what the content-neutral insert
does. 
$p_0 = 0.5$ stands on a structural ground: moving it would recalibrate a
preregistered test against a different manipulation. 
The five resolving cells license no claim about direction. 
The statistic is reported in full and the reading of its sign is what is
blocked. No verdict on the preregistered hypothesis about adversary tracking is
issued. 

\subsection{Why the direction question is unanswerable here}
\paragraph{The structural reason}
Quantity (c) asks whether choices moved toward $A = 1$. On every confirmatory
item, $A = 1$ is both the adversary-aware optimum and the salience pole
(Section~\ref{sec:coincidence}). No function of a chosen option can separate
two labels attached to the same option, so no quantity computed on this set,
and no successor measure on this tile, can support a claim that a model does or
does not respond to the adversary. The separation is absent from the item set,
not from the coordinate. The tile is not revisitable: Paper 1 fixed
\texttt{size} as the primary tile on measured grounds before any of this was
known, and the frozen set's eight separating items are on the two tiles Paper 1
found carry approximately nothing. This is why the direction question was never
answerable on this set. It is not a null, and it is not a caveat on a result.

\paragraph{Three independent premises reach the same place}
Before the structural reason was recorded, the design had already refused a
direction reading on three grounds, each sufficient alone and each still true.
First, a direction claim was preregistered to rest on excess over the model's
marginal null, and none of the three quantities is that excess. Second, the
content-neutral insertion's own departure resolves below 0.5 on CTRL, so a
departure below 0.5 is not established as a property of adversary content
rather than of inserted text. Third, the sign test discards moves between
options that share an $A$, and on three of the five resolving cells, CTRL under
$F_1$ and L3 under both framings, those discarded switches are concentrated at
1.81 to 4.09 times the rate the per-pair bound assumes. 
That is selection, not measurement error: nothing establishes that the
discarded switches carry the direction of the retained ones. The structural
reason explains the three from the other side.

\paragraph{The preregistered defence has no successor}
The preregistration defended against a generic, prompt-induced shift in option
preference with an attribution cap: reweight the confirmatory set by the option
shift measured on the control set and require the observed mean change in $A$
to exceed it. We computed the cap and the marginal null as specified. Both are
descriptive, both inherit the mixture defect of Section~\ref{sec:discussion},
and no mapping from either onto the sign test exists or may be written. 
Arm B's confirmatory family therefore carries no defence against the
generic-shift hypothesis and cannot acquire one. That is a finding about the
design, and it is the same finding as the coincidence: a defence against
generic movement is a defence of a direction claim, and there is no direction
claim to defend.

\paragraph{What survives}
The movement claim. The adversary framings changed the chosen option on every
model, on 30 to 77 of 108 items, 
against an exact no-effect rate. That needs no direction and is untouched by
the coincidence. What this evaluation cannot say is what the movement was
toward, and the reason is on the item set, fixed before any model was run.

\section{Discussion and Limitations}\label{sec:discussion}
Each limitation below is a property of the item set or the coordinate, not of
any model, and none is repaired by anything the design can do.

\paragraph{$n = 108$}
The confirmatory set holds 108 items against a preregistered benchmark of 400. 
The redraw that would have enlarged it was declined, because sizing it needed a
quantity not observable before the arm ran. At the realized effective $n$, the
sign test's power at an alternative of 0.75 runs from 0.25 to 0.91 across
cells. 
A modest directional effect would not be detected, and the test has least
power exactly where no movement is the truth, because ties carry no sign. 
Under the coincidence this limit is moot for direction, but it bounds every
per-cell figure in Table~\ref{tab:armb}.

\paragraph{The coordinate cannot see some moves}
Two options can share a value of $A$, and a switch between them is invisible to
the sign test. On the confirmatory set 84 of 1{,}620 option pairs are tied on
$A$, 5.2 percent, against 3.3 percent pooled over the divergence set. 
Per item, 73 of 108 contain at least one tied pair among their fifteen, a figure
inflated by option count and reported only beside the per-pair one. 
\texttt{size} is the worst tile on this measure, against 2.3, 1.3 and 0.0
percent on \texttt{manmade}, \texttt{hold} and \texttt{moves}. 
The tie tolerance is Paper 1's $10^{-12}$, 
and the next gap above it is 0.0020, 
so any tolerance in that range classifies identically. Realized switches do
not respect the per-pair bound: the concentration ratio runs above 1 on some
cells and below on others, and the largest, 5.10, is on CTRL under $F_2$, a
cell that does not resolve. 
The concentrated switches fall mostly on canonical options 4 and 5. 
Those are option ids, not menu positions, since Paper 1's rendering permutes
the menu, so this is not evidence of a position preference. 

\paragraph{Why the mean is descriptive}
$A$ divides by the item's adversary extent, so an option well below
$\ostar{0}$ on normalized margin is far negative. Split by Paper 1's
salience-to-Bayes poles, the 1{,}750 option cells of the divergence set are
470 at the Bayes pole with median $A$ of 0, 439 at the salience pole with
median $A$ of 1, and 841 off-pole with median $A$ of $-3.33$. 
Split by $A$'s own value the counts are 464 at exactly 0, 439 at exactly 1,
and 847 elsewhere. 
Of 6{,}048 admissible ordered option pairs, 70.0 percent change $A$ by more
than the target move of 1 in magnitude and 15.2 percent by more than 10, 
so one item moving to a far off-pole option outweighs ten items making the
exact move the arm exists to detect. This is a measurement of the coordinate,
not of any model. 
It is why the confirmatory instrument is a change rate plus a sign test. For
the same reason we do not report a negative mean $A$ as a finding: it restates
Paper 1's far-side result on the margin coordinate. 

\paragraph{The neutral baseline}
The content-neutral insertion's sign proportion (Section~\ref{sec:armb}) is
the only direction-matched reference the design has, and it resolves on one
model. On the content-neutral contrast the coordinate's blind spot, the tie
rate minus the same-option rate, runs from 0 to 0.051. 

\paragraph{No human arm}
Paper 2 collects no human data. The coincidence is a property of the signal
space and holds for any chooser, but whether people who are told about the
adversary move toward the salience pole, or away from it, is not measured here.

\paragraph{A learned policy is not run}
A trained speaker would need an item set on which the adversary-aware optimum
and the salience pole are different options, and a coordinate defined on those
items. The frozen set supplies the first on 8 items and the second on none of
them, 
and the pool repeats the pattern at scale (Section~\ref{sec:coincidence}).
Which item properties would produce a separating item with a defined coordinate
is not characterized, and we do not propose an item set.

\paragraph{The check, and its scope}
The coincidence was visible in model-free geometry: the confirmatory count
needs only $\ostar{\infty}$ and $\ofit$ for each item. On this task it was the
check that decided what the evaluation could say. We claim no more than that it
was necessary here and that it is cheap to run; we do not claim that robust
and heuristic targets coincide in other tasks.

\section{Conclusion}
An informed adversary sharing the audience changes the optimal signal in
Paper 1's task, exactly and computably: on 18.2 percent of the candidate
pool 
some finite budget moves the optimum, and at $\beta = 0$, under the inherited
$\tau = 1$, the game is Paper 1's. Where the optimum moves, it moves to the
salience pole. On all 108 confirmatory items the adversary-robust signal is 
the one that fits the target best, and across the pool the two separate only
where Paper 1's coordinate is undefined. The same fact bounds the evaluation.
Adversary framings changed seven models' choices on 30 to 77 of 108 items, 
and on these items no measurement can say whether that movement was toward the
adversary-aware target or toward salience, because they are one option. The
check that exposed this reads no model output, and it belongs before any
evaluation of adversary-awareness, not after.

\section*{Data Availability}
Released in the accompanying repository: the adversary specification, the
decision log and every preregistration version, the emitted result artifacts
every number above is read from, and the code that produced them. Paper 1's frozen items, renderings and choices are used
unchanged and are released with Paper 1.

\section*{Ethics Statement}
No human participants were recruited and no participant data is reported.

\appendices
\section{Derivations and the $\beta_c$ Cross-Check}\label{app:derivations}
\paragraph{R1}
With $L_h = L(h\mid o)$ for fixed $o$, $P(\text{choose } h\mid o,d) =
L_h^{1/\tau} e^{(\beta/\tau)\mathbb{1}[h=d]} / Z(d)$, where
$Z(d) = \sum_{h'} L_{h'}^{1/\tau} + L_d^{1/\tau}(e^{\beta/\tau} - 1)$.
For $h = h^*$ and $d \ne h^*$ the numerator does not depend on $d$ and the
denominator is increasing in $L_d$, since $e^{\beta/\tau} - 1 \ge 0$. The
adversary minimizes by maximizing $L_d$, giving
$d^*(o) = \arg\max_{h\ne h^*} L(h\mid o)$ for every $\tau > 0$, $\beta \ge 0$.

\paragraph{R2}
Substituting $d^*(o)$ gives Eq.~(4). At $\beta = 0$ the second term vanishes.
Writing $w(o,h) = f(o,h)/\sum_{o'} f(o',h)$, $L(h\mid o) = w(o,h)/\sum_{h'}w(o,h')$,
so $\sum_h L(h\mid o) = 1$ for every $o$ and every strictly positive $f$, and at
$\tau = 1$, $V_0(o) = L(h^*\mid o)$. At $\tau \ne 1$ two options with equal
$L(h^*\mid o)$ and differently peaked rival distributions have different
$\sum_h L(h\mid o)^{1/\tau}$, so $V_0$ reorders them. As $\beta \to \infty$,
$V_\beta(o) \sim e^{-\beta/\tau}[L(h^*\mid o)/M(o)]^{1/\tau}$, and
$x \mapsto x^{1/\tau}$ is increasing, so the argmax tends to the margin argmax
at every $\tau$.

\paragraph{Single crossing}
With $C_o = L(h^*\mid o)^{1/\tau}$, $B_o = M(o)^{1/\tau}$ and
$A_o = \sum_h L(h\mid o)^{1/\tau} - B_o$, $V_\beta(o) = C_o/(A_o + B_o x)$.
Equating two options gives
$x\,(C_o B_{o'} - C_{o'} B_o) = C_{o'} A_o - C_o A_{o'}$, linear in $x$, with
at most one root unless the coefficient of $x$ is zero. Since $\ostar{0}$
leads at $x = 1$ by definition and tie-break, each rival overtakes at most once
and never returns, so $D(\cdot)$ is nested and bisection on the indicator
$\mathbb{1}[\arg\max_o V_\beta(o) \ne \ostar{0}]$ converges to $\beta_c$.

\paragraph{The divergence identity}
$\beta_c$ is finite exactly when $\ostar{0} \ne \ostar{\infty}$ under the
tie-broken argmax. 
The same identity read off the raw sign of the margin headroom fails on one
item, at a headroom of $4.4\times10^{-16}$, 
which is why the divergence set is taken from the finite $\beta_c$.

\paragraph{Closed form against bisection}
On the frozen 1{,}000 the two methods classify every item identically, 
with the largest gap in $\beta$ at $2.2\times10^{-6}$. 
On the pool, an exact test for a zero coefficient of $x$ disagreed with
bisection on 161 items. 
Direct evaluation of $V_\beta$ settles each: on all 62 where the closed form
found a root and bisection did not, both coefficients are rounding residue on
numerically identical curves and $\ostar{0}$ never loses, 
and where bisection found a crossing, $\ostar{0}$ never regains the lead. 
Bisection's own agreement with the divergence identity holds by construction
and is not evidence. 
Reading the coefficient of $x$ as zero at magnitude $10^{-12}$, the spec's
inherited absolute tolerance, 
which sits inside an empty gap in the pool's coefficients from
$3.5\times10^{-17}$ to $7.3\times10^{-8}$, 
the closed form disagrees with bisection on 136 items, 
exactly the tolerance-determined set of Appendix~\ref{app:tolerance}. 
That residual is expected: the closed form tests strict crossing and cannot see
a flip made by the tie-break.

\section{Tolerance-Determined Divergent Items}\label{app:tolerance}
Of the pool's 36{,}464 divergent items, 136 have a $\beta_c$ set by Paper 1's
tie-break band rather than by a strict crossing, 
and all 136 are divergent only by tie-break: no rival ever leads $\ostar{0}$
beyond the band at any $\beta$. 
Every one has $\beta_c$ between 17.35 and 23.45, 
above the reporting grid's endpoint of 8, 
so $|D(8)|$ and every grid-rate figure are unaffected and only $|D(\infty)|$ 
includes them. None of the 2{,}748 separating items is among them, 
and Paper 1's frozen 1{,}000 contains none, so no confirmatory figure is
touched. 
They stay in $D(\infty)$ because the tie-broken argmax is the preregistered
definition of divergence, and excluding them would redefine divergence after
seeing where it bites.

\section{Arm A: Alternative Denominators and Exploratory Analysis}\label{app:arma}
\paragraph{Other denominators for the price}
Neither is the headline. Measured against the posterior mass the Bayes-optimal
signal earns above the uniform prior of 0.01, the mean of per-item ratios is
0.578; 
42 items exceed 1 there, where the robust signal leaves the truth below the
prior, and are kept. 
As a share of total posterior mass it is 0.182. 
Paper 1 normalizes by the item's posterior range, not by the prior, which is
why $\postnorm$ is primary. The nineteen items at exactly 1 have mean
\fitcost{} 0.386 against 0.195 for the rest. 
No divergent item has a price of exactly zero, 2 are Paper 1 boundary-exact
items, and the price on robust items is 0. 

\paragraph{Conflict enrichment}
Paper 1's frozen set is 50 percent conflict items against 18.7 percent of the
pool, 
and conflict items are 98.3 and 92.4 percent of the two divergence sets. 
The frozen price on $\postnorm$ exceeds the pool's by 0.029. 
This is a descriptive robustness check, not a second estimate. 

\paragraph{What predicts a low $\beta_c$ (exploratory)}
There is no preregistered prediction and no threshold, and no coefficient below
is a test. On the 460 finite items, Spearman correlations of $\beta_c$ are
$-0.779$ with margin headroom, $+0.602$ with Paper 1's posterior gap, $-0.574$
with \fitcost{} and $-0.557$ with Paper 1's decision margin. 
Margin headroom restates divergence among robust items and is used only for
magnitude among finite ones. 
An OLS model of $\log\beta_c$ on standardized item properties and tile reaches
$R^2 = 0.656$ on 460 items; a negative coefficient means a cheaper
adversary. 
Neither option count nor the fraction of distinct fit vectors reproduces the
tiles' order of existence rates. 
With four tiles a rank statistic takes five values and separates nothing, so
this is an ordering check, not evidence. 

\section{Full Per-Cell Tables}\label{app:tables}
Table~\ref{tab:full} gives the intervals behind Table~\ref{tab:armb}, with the
control's two measures on their bases.

\begin{table*}[t]
\centering
\caption{Intervals and control measures per cell, \texttt{size} tile, PMI rule. (a) rate: item-mean change rate on 108 items with bootstrap interval (10{,}000 item resamples, seed 20260910, level $1-0.05/21$). Control rate: the same on the 142 control items, exploratory; not comparable to (a) as an attribution. TV: total variation between the framing and $F_0$ option marginals on the 142 control items and, beside it, the item-weighted within-tile mean over all 540 control items. (b): interval on the difference from $R_m$. (c): 95\% interval on the sign proportion, and realized power at an alternative of 0.75.}
\label{tab:full}
\resizebox{\textwidth}{!}{%
\begin{tabular}{llccccccc}
\toprule
model & framing & (a) rate [interval] & control rate [interval] & TV 142 & TV 540 & (b) interval & (c) 95\% interval & power at 0.75 \\
\midrule
CTRL & $F_1$ & 0.231 [0.148, 0.325] & 0.268 [0.183, 0.363] & 0.173 & 0.081 & [$-0.014$, $+0.144$] & [0.064, 0.328] & 0.44 \\
B2   & $F_1$ & 0.264 [0.181, 0.352] & 0.243 [0.165, 0.327] & 0.056 & 0.097 & [$-0.162$, $+0.032$] & [0.300, 0.587] & 0.64 \\
B4   & $F_1$ & 0.394 [0.292, 0.505] & 0.437 [0.345, 0.525] & 0.158 & 0.162 & [$-0.134$, $+0.083$] & [0.351, 0.605] & 0.83 \\
L1   & $F_1$ & 0.162 [0.088, 0.245] & 0.165 [0.106, 0.232] & 0.081 & 0.168 & [$-0.112$, $+0.069$] & [0.126, 0.511] & 0.25 \\
L2   & $F_1$ & 0.208 [0.134, 0.292] & 0.162 [0.099, 0.232] & 0.042 & 0.055 & [$-0.153$, $+0.028$] & [0.364, 0.680] & 0.51 \\
L3   & $F_1$ & 0.296 [0.204, 0.394] & 0.271 [0.190, 0.356] & 0.187 & 0.151 & [$-0.195$, $-0.019$] & [0.091, 0.333] & 0.61 \\
L4   & $F_1$ & 0.259 [0.167, 0.356] & 0.317 [0.229, 0.402] & 0.215 & 0.324 & [$-0.088$, $+0.097$] & [0.109, 0.364] & 0.64 \\
\midrule
CTRL & $F_2$ & 0.250 [0.167, 0.333] & 0.239 [0.162, 0.324] & 0.046 & 0.052 & [$-0.042$, $+0.144$] & [0.208, 0.538] & 0.44 \\
B2   & $F_2$ & 0.347 [0.255, 0.444] & 0.306 [0.225, 0.394] & 0.070 & 0.154 & [$-0.245$, $-0.051$] & [0.223, 0.470] & 0.81 \\
B4   & $F_2$ & 0.472 [0.370, 0.574] & 0.472 [0.387, 0.560] & 0.116 & 0.234 & [$-0.231$, $+0.024$] & [0.356, 0.593] & 0.91 \\
L1   & $F_2$ & 0.171 [0.097, 0.255] & 0.173 [0.109, 0.239] & 0.102 & 0.166 & [$-0.130$, $+0.061$] & [0.107, 0.449] & 0.26 \\
L2   & $F_2$ & 0.190 [0.120, 0.269] & 0.155 [0.095, 0.222] & 0.018 & 0.025 & [$-0.130$, $+0.042$] & [0.338, 0.662] & 0.58 \\
L3   & $F_2$ & 0.352 [0.255, 0.454] & 0.303 [0.222, 0.384] & 0.187 & 0.169 & [$-0.245$, $-0.079$] & [0.141, 0.378] & 0.76 \\
L4   & $F_2$ & 0.356 [0.255, 0.459] & 0.370 [0.282, 0.462] & 0.278 & 0.300 & [$-0.199$, $+0.009$] & [0.119, 0.337] & 0.83 \\
\bottomrule
\end{tabular}}
\end{table*}

The TV of record never exceeds the control change rate on the thirteen cells
where both were formed on identical renderings; on B2 under $F_1$ that premise
does not hold. 
No marginal is pooled across tiles of different arity. 
The marginal null and the attribution cap are emitted per model and framing in
the released artifacts; both are descriptive and neither is read against
quantity (c). 

\section{Preregistration and Deviations}\label{app:prereg}
Paper 2's preregistration is version 2.0, amended in versions 2.1 through 2.22,
each a new file; no version is edited in place. Every decision after version
2.0 is logged with its alternatives in a numbered decision log (P2-D1 to
P2-D29). The material deviations from version 2.0 follow.

\paragraph{The \fitcost{} gate fired and Paper 1's items are reused (P2-D3, P2-D4, P2-D7)}
Version 2.0 planned a fresh $\beta_c$-stratified draw. The gate on $\beta_c$
tracking \fitcost{} fired (Section~\ref{sec:arma}), so Arm B runs on the
\texttt{size} tile of Paper 1's frozen 1{,}000, restricted to finite $\beta_c$.
The unit of analysis is per tile, \texttt{size} only; pooling across tiles was
declined because Paper 1 measured the other tiles carrying approximately
nothing. An enlargement of \texttt{size} alone was declined rather than
deferred, because sizing it needed an unmeasured quantity. The cost is
$n = 108$ against a benchmark of 400. 

\paragraph{The mean is replaced by a sign test (P2-D6, P2-D10)}
Version 2.0's confirmatory test was a paired test on mean $\Delta A$. The
three-way mixture on $A$ (Section~\ref{sec:discussion}) makes the mean's
expectation a different quantity from the effect, so the confirmatory
instrument became a same-option rate plus an exact sign test against 0.5 at
$0.05/21$, with the mean kept as a descriptive quantity. 

\paragraph{The reference gate is removed (P2-D8, P2-D9, P2-D12, P2-D13)}
An intermediate version gated the movement half on the framing moving choices
more than Paper 1's condition-5 sentence. Once the no-effect rate was shown to
be exactly 1, that gate was removed: three quantities are reported and none
gates another, with the condition-5 comparison kept as descriptive context and
a numerical noise floor of 7 items on quantity (a). 

\paragraph{The unit of the sign test is corrected to the item (P2-D20)}
Three readings of quantity (c)'s unit were in circulation. The governing
documents both said item, and the item unit was restored, with a rule for items
whose two permutations disagree in sign.

\paragraph{$p_0 = 0.5$ is retained on the structural ground (P2-D14, P2-D21, P2-D22)}
An earlier justification of $p_0$ by the content-neutral baseline was withdrawn
when, at the item unit, B2's baseline point estimate sat above 0.5. $p_0$ stands
on the ground that moving it recalibrates a preregistered test.

\paragraph{No direction claim, and the tracking question is closed (P2-D5, P2-D24, P2-D25, P2-D26)}
Version 2.0 predicted that models would not track the adversary and
preregistered a test of it. P2-D5 recorded the divergence-set coincidence and
required any content claim to rest on excess over the marginal null. P2-D24
ruled that the five resolving cells license no direction claim; P2-D25 made
the marginal null and attribution cap descriptive, with no successor; P2-D26
discharged the question as unanswerable on the confirmatory set, on the
structural ground of Section~\ref{sec:coincidence}. No verdict on the
preregistered hypothesis is issued.

\paragraph{The control change rate is authorized as exploratory (P2-D27)}
Version 2.0's control measure is total variation. A change rate on the control
set, in quantity (a)'s units, is a new quantity, authorized after (a) was
published as exploratory and side by side only, with no difference, ratio or
share against (a).

\paragraph{Tolerance-determined items and the closed-form tolerance (P2-D28, P2-D29)}
The tie-broken definition of divergence is kept and the 136 items it admits 
by tie-break are disclosed (Appendix~\ref{app:tolerance}). The closed-form
cross-check's zero test was set inside an empty gap at the inherited tolerance.

\paragraph{Not run}
Version 2.0 preregistered a third arm, a learned speaker policy. It is not run:
the author ruled it out on the counting result of Section~\ref{sec:coincidence},
recorded without a decision number. The preregistered robustness check at
$\tau \in \{0.5, 2\}$ has not been run and nothing here depends on it. No
result of version 2.0's decay-family sensitivity analysis or of its
three-scoring-rule recomputation is among the artifacts this paper reports;
every Arm B figure is under the PMI rule.

\section{Item Construction}\label{app:items}
Items are Paper 1's, unchanged. Paper 1 generated 200{,}000 candidate items 
from a vocabulary of 1{,}118 concepts, 
one tile per item, and selected a frozen set of 1{,}000, 
250 per tile, 
half conflict and half control. 
The frozen set's divergent items number 114, 122, 116 and 108 on
\texttt{manmade}, \texttt{moves}, \texttt{hold} and \texttt{size}. 
Arm B's confirmatory set is the 108 on \texttt{size}; its control set is the
remaining 142 \texttt{size} items, 
and the 540 adversary-robust items across all four tiles are reported beside
it. 
The adversary-extent floor of 0.02 on $1 - \textit{mn}_i(\ostar{0})$ applies
only to aggregates of per-item $A$ and leaves 101 confirmatory items there. 
Each item has two stored option-order permutations, and every prompt is
Paper 1's Format V rendering.

\section{The Three Framings, Verbatim}\label{app:framings}
$F_0$ is Paper 1's Format V prompt. Braces mark per-item content.
\begin{verbatim}
Two lists of objects are shown below.

LEFT
{left}

RIGHT
{right}

One object from LEFT and one from
RIGHT have been secretly chosen as a
pair. You know the pair: {tm} and {tc}.

Another person can see both lists but
does not know the pair. Your job is to
help them work out which pair it is.

You may send exactly one of the
following descriptions, and nothing
else:

{options}

Which do you send?
\end{verbatim}
Lines are rewrapped here for the column; the prompt's own line breaks are
Paper 1's. $F_1$ and $F_2$ insert one block, as its own paragraph, immediately
before \texttt{Which do you send?}, and change nothing else.

\noindent $F_1$:
\begin{verbatim}
A third person also sees both lists
and already knows the pair. After you
send your description, that person
will speak to the other person, and
wants them to name a different pair.
They know the pair, and the other
person still does not know what it is.
\end{verbatim}

\noindent $F_2$:
\begin{verbatim}
A third person also sees both lists
and already knows the pair. After you
send your description, that person
will speak to the other person, and
wants them to name a different pair.
They will argue for whichever wrong
pair your description makes look most
likely.
\end{verbatim}
The two blocks share their first two sentences verbatim; only the third
differs. $F_1$'s third sentence restates two facts the prompt already gives, so
the $F_1$ against $F_2$ contrast is the adversary's rule and not length.

\bibliographystyle{IEEEtran}
\bibliography{references}

\begin{thebibliography}{10}
\providecommand{\url}[1]{#1}
\csname url@samestyle\endcsname
\providecommand{\newblock}{\relax}
\providecommand{\bibinfo}[2]{#2}
\providecommand{\BIBentrySTDinterwordspacing}{\spaceskip=0pt\relax}
\providecommand{\BIBentryALTinterwordstretchfactor}{4}
\providecommand{\BIBentryALTinterwordspacing}{\spaceskip=\fontdimen2\font plus
\BIBentryALTinterwordstretchfactor\fontdimen3\font minus \fontdimen4\font\relax}
\providecommand{\BIBforeignlanguage}[2]{{%
\expandafter\ifx\csname l@#1\endcsname\relax
\typeout{** WARNING: IEEEtran.bst: No hyphenation pattern has been}%
\typeout{** loaded for the language `#1'. Using the pattern for}%
\typeout{** the default language instead.}%
\else
\language=\csname l@#1\endcsname
\fi
#2}}
\providecommand{\BIBdecl}{\relax}
\BIBdecl

\bibitem{huynh2026consistency}
C.~Huynh, ``Consistency without alignment: Item-sensitive language models indistinguishable from random,'' arXiv:2609.00576, 2026.

\bibitem{zheng2025stackelberg}
Z.~Zheng, Ye, Zhao, and Wang, ``The {Stackelberg} speaker: Optimizing persuasive communication in social deduction games,'' arXiv:2510.09087, 2025.

\bibitem{stephenson2024codenames}
Stephenson, Sidji, and Ronval, ``Codenames as a benchmark for large language models,'' arXiv:2412.11373, 2024.

\bibitem{bills2024cooperation}
Bills, Archibald, and Blaylock, ``Improving cooperation in language games with {Bayesian} inference and the cognitive hierarchy,'' arXiv:2412.12409, 2024.

\bibitem{siu2022codenames}
Siu, ``Towards automating {Codenames} spymasters with deep reinforcement learning,'' arXiv:2212.14104, 2022.

\bibitem{kamenica2011persuasion}
Kamenica and Gentzkow, ``{Bayesian} persuasion,'' \emph{American Economic Review}, vol. 101, no.~6, pp. 2590--2615, 2011.

\bibitem{gentzkow2017competition}
Gentzkow and Kamenica, ``Competition in persuasion,'' \emph{Review of Economic Studies}, vol.~84, no.~1, pp. 300--322, 2017.

\bibitem{gentzkow2017rich}
------, ``{Bayesian} persuasion with multiple senders and rich signal spaces,'' \emph{Games and Economic Behavior}, vol. 104, pp. 411--429, 2017.

\bibitem{ravindran2020zerosum}
Ravindran and Cui, ``Competing persuaders in zero-sum games,'' arXiv:2008.08517, 2020.

\bibitem{hossain2024multisender}
Hossain, Wang, Lin, Chen, Parkes, and Xu, ``Multi-sender persuasion: A computational perspective,'' arXiv:2402.04971, 2024.

\bibitem{wang2026mindgames}
Wang \emph{et~al.}, ``{MINDGAMES}: A live arena for evaluating social and strategic reasoning in multi-agent {LLMs},'' arXiv:2605.29512, 2026.

\bibitem{yuan2026quack}
Yuan \emph{et~al.}, ``{QUACK}: Questioning, understanding, and auditing communicated knowledge in multimodal social deduction agents,'' arXiv:2605.27068, 2026.

\bibitem{hazan2011acoustic}
Hazan and Baker, ``Acoustic-phonetic characteristics of speech produced with communicative intent to counter adverse listening conditions,'' \emph{Journal of the Acoustical Society of America}, vol. 130, no.~4, pp. 2139--2152, 2011.

\bibitem{buz2016dynamically}
Buz, Tanenhaus, and Jaeger, ``Dynamically adapted context-specific hyper-articulation: Feedback from interlocutors affects speakers' subsequent pronunciations,'' \emph{Journal of Memory and Language}, vol.~89, pp. 68--86, 2016.

\bibitem{gessa2026louder}
Gessa, Valzolgher, Giovanelli \emph{et~al.}, ``Will {I} speak louder if {I} see you struggling to understand? speech modifications in response to non-verbal visual cues of listening effort,'' \emph{Psychonomic Bulletin and Review}, 2026, doi:10.3758/s13423-026-02942-3.

\bibitem{rode2010truth}
Rode, ``Truth and trust in communication: Experiments on the effect of a competitive context,'' \emph{Games and Economic Behavior}, vol.~68, no.~1, pp. 325--338, 2010.

\bibitem{sutter2009deception}
Sutter, ``Deception through telling the truth?! experimental evidence from individuals and teams,'' \emph{Economic Journal}, vol. 119, no. 534, pp. 47--60, 2009.

\bibitem{gneezy2005deception}
Gneezy, ``Deception: The role of consequences,'' \emph{American Economic Review}, vol.~95, no.~1, pp. 384--394, 2005.

\bibitem{cai2006overcommunication}
Cai and Wang, ``Overcommunication in strategic information transmission games,'' \emph{Games and Economic Behavior}, vol.~56, no.~1, pp. 7--36, 2006.

\bibitem{pinker2008indirect}
Pinker, Nowak, and Lee, ``The logic of indirect speech,'' \emph{Proceedings of the National Academy of Sciences}, vol. 105, no.~3, pp. 833--838, 2008.

\end{thebibliography}
\end{document}